\pdfoutput=1

\documentclass[11pt]{article}
\usepackage[]{acl}

\usepackage{amsmath}
\usepackage{times}
\usepackage{latexsym}
\usepackage{hyperref}
\usepackage[T1]{fontenc}
\usepackage{float}
\usepackage{stfloats}
\usepackage{graphicx} 
\usepackage{footnote}
\usepackage[utf8]{inputenc}
\usepackage{microtype}
\usepackage{algorithm}
\usepackage{algorithmic}
\usepackage{inconsolata}
\usepackage{multirow}
\usepackage{graphicx}
\usepackage{listings}
\usepackage{xcolor}
\usepackage{booktabs}
\usepackage{adjustbox}
\usepackage{amsmath}
\usepackage{setspace}
\usepackage{array,caption}
\usepackage[labelfont=bf]{caption}
\usepackage{longtable}

\usepackage{times}
\usepackage{latexsym}

\usepackage[T1]{fontenc}

\usepackage[utf8]{inputenc}

\usepackage{microtype}

\usepackage{inconsolata}

\title{RepCali: High Efficient Fine-tuning Via Representation Calibration in Latent Space for Pre-trained Language Models}

\author{ Fujun Zhang \\
  College of Computer  \\ Science, \\ Inner Mongolia \\ University,\\
  \texttt{zfjimu@163.com} \\\And
  Xiaoying Fan \\
  College of Computer \\ Science, \\ Inner Mongolia \\ University,\\
  \texttt{32209144@mail.imu.edu.cn} \\\And
  XiangDong Su \\
  College of Computer \\ Science, \\ Inner Mongolia \\ University, \\
  \texttt{cssxd@imu.edu.cn} \\\And
  Guanglai Gao \\
  College of Computer \\ Science, \\ Inner Mongolia \\ University, \\
  \texttt{ccsggl@imu.edu.cn}
  \\}

\begin{document}
\maketitle
\begin{abstract}
Fine-tuning pre-trained language models (PLMs) has become a dominant paradigm in applying PLMs to downstream tasks. However, with limited fine-tuning, PLMs still struggle with the discrepancies between the representation obtained from the PLMs' encoder and the optimal input to the PLMs' decoder. This paper tackles this challenge by learning to calibrate the representation of PLMs in the latent space. In the proposed representation calibration method (RepCali), we integrate a specific calibration block to the latent space after the encoder and use the calibrated output as the decoder input. The merits of the proposed RepCali include its universality to all PLMs with encoder-decoder architectures, its plug-and-play nature, and ease of implementation.  Extensive experiments on 25 PLM-based models across 8 tasks (including both English and Chinese datasets) demonstrate that the proposed RepCali offers desirable enhancements to PLMs (including LLMs) and significantly improves the performance of downstream tasks. Comparison experiments across 4
benchmark tasks indicate that RepCali is superior to the representative fine-tuning baselines.
\end{abstract}

\section{Introduction}
Pre-trained language models (PLMs) exhibit impressive capabilities in capturing both syntactic and semantic information within text data, rendering them highly valuable for a range of downstream tasks \cite{DBLP:conf/naacl/DevlinCLT19}. 
In reality, the pre-training data is usually domain-general while the downstream task data is significantly varied with domains, and the targets of the pre-training tasks and the downstream tasks are quite different. 
Due to the domain gaps and objective gaps between the pre-training tasks and the downstream tasks, when applied to specific downstream tasks, the PLMs need to be trained with task-specific data to enhance their ability to process the language features relevant to that task. Therefore, fine-tuning of the PLMs has become a dominant paradigm in applying PLMs to downstream tasks~\cite{ruder2021lmfine-tuning}.

\begin{figure}[t]
\scalebox{1}[1]{
\centering
\includegraphics[width=0.95\columnwidth]{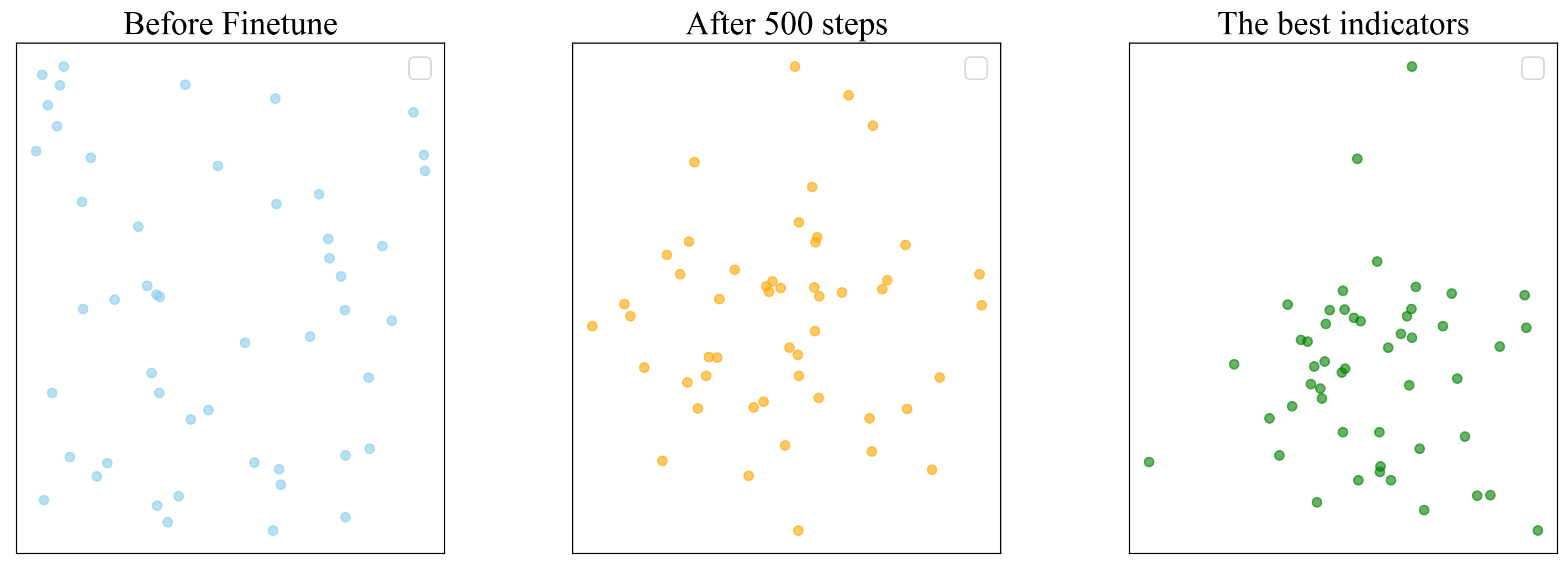}
}
\caption{Visualization of the distribution of the output of T5 encoder before and after fine-tuning via the T-SNE algorithm on the Abductive Commonsense Reasoning ($\alpha$NLG) task.} 

\label{fig:example}
\end{figure}

As shown in Figure \ref{fig:example}, in the latent space characterization analysis, we observe that the output of the PLMs(T5) encoder before fine-tuning exhibits a disordered distribution, while its distribution structure is significantly compacted after fine-tuning, but has not yet reached the optimal state. This suggests that although the fine-tuning can effectively improve downstream task performance, PLMs are still difficult to be fully adapted to the target domain properties within a limited number of fine-tuning cycles ~\cite{ruder2021lmfine-tuning, DBLP:conf/nips/ChenJWW0L22}. \textbf{In essence, the current performance bottleneck stems from the fact that there is still a non-negligible inter-domain discrepancy between the distribution of the model encoder's representations in the target domain latent space and the optimal input distribution expected by the decoder}.


\citet{DBLP:conf/www/0001ZHZ022} learns and clusters in the embedding latent space in the PLM to improve the diversity and quality of model generation. \citet{DBLP:conf/emnlp/LiGLPLZG20} confirms the importance of learning latent space. 
Based on this, we argue that it will be more effective to directly adjust the representation from the PLMs' encoder in latent space through a learnable block in the fine-tuning process. Therefore, this paper proposes RepCali, a simple and effective representation calibration method that integrates a well-designed calibration block to the latent space after the PLMs' encoder and uses the calibrated output as the input of the PLMs' decoder for PLM fine-tuning. The calibration block only involves shape seed, learnable embedding and layer normalization.

\begin{figure*}[t]
\centering
\scalebox{0.99}{
\includegraphics[width=1.85\columnwidth]{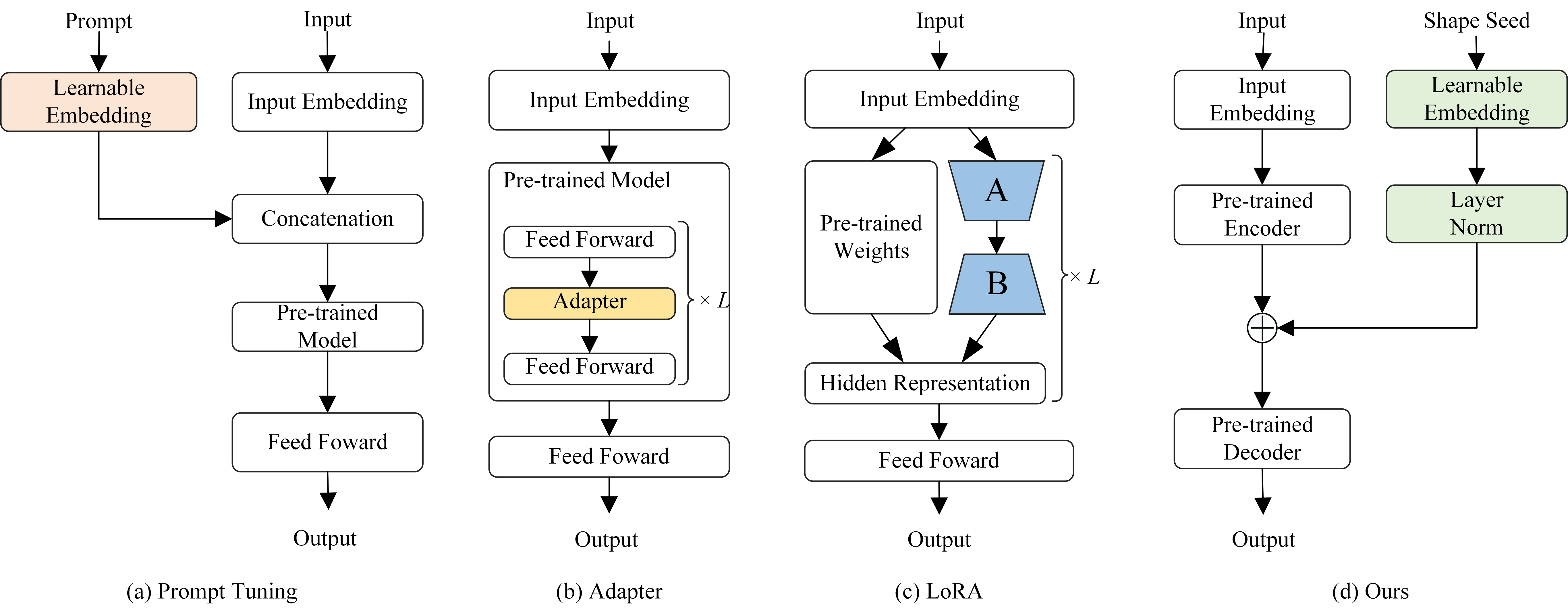}
}
\caption{The detailed architecture of various tuning methods.}
\label{fig:compared}
\end{figure*}

\textbf{It is worth noting that our representation calibration method differs from other fine-tuning methods like prompt tuning, adapter, and LoRA, as shown in Figure \ref{fig:compared}}. Prompt tuning usually contains learnable parameters and appends the learned prompt embedding to the input embedding to guide the pre-trained models. Adapter layers are small neural network modules inserted between the layers of a pre-trained model and their parameters are updated during fine-tuning. LoRA introduces low-rank matrices to modify the self-attention mechanism of transformers and updates only these low-rank matrices. Different from these methods, our method introduces a specialized representation calibration block between the PLM's encoder and decoder, which calibrates the encoder output before it is fed into the decoder. Hence, the PLM's decoder receives an improved input and generates a better result.

Extensive experiments on 25 PLM-based models across other 8 NLP downstream tasks demonstrate that RepCali significantly enhances the PLMs (including large language models (LLMs)) yielding substantial improvements for PLMs.  Comparison experiments with 4 representative fine-tuning baselines across 3 benchmark tasks indicate that our proposed fine-tuning method, RepCali, is superior to these baselines. The merits of the proposed representation calibration method RepCali include its universal applicability to all PLMs with encoder-decoder architectures, its plug-and-play nature and its ease of implementation, with only a marginal increase in model parameters.  Our experiments include both English and Chinese datasets, and the results both show that RepCali generalizes effectively to different languages.

\begin{figure*}[ht]
\centering
\scalebox{0.98}{
\includegraphics[width=1.98\columnwidth]{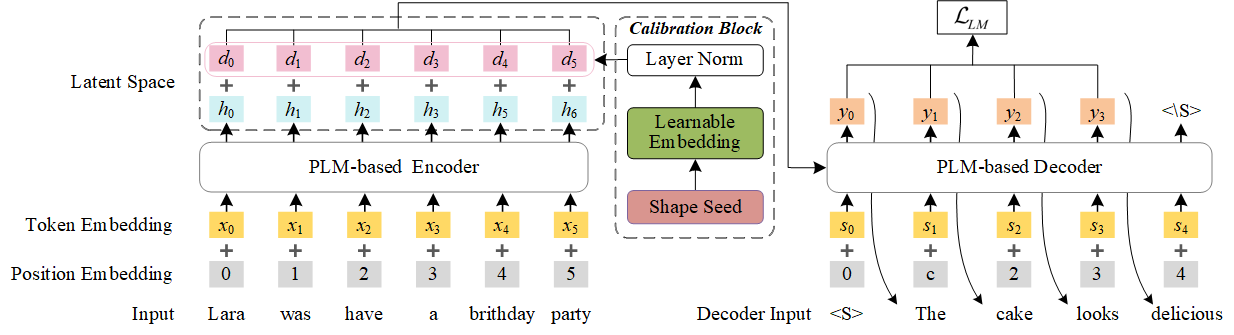}
}
\caption{Overview of our representation calibration method. }
\label{fig:model}
\end{figure*}

\section{Related Work}

\subsection{Fine-tuning Method}
In recent years, adapters have gradually become mainstream in PLM fine-tuning.
\citet{DBLP:conf/icml/HoulsbyGJMLGAG19} proposed an adapter module that introduces a minimal number of trainable parameters for each task, enabling the addition of new tasks without affecting previously trained ones.
\citet{ruckle2020adapterdrop} presented AdapterDrop, which strategically eliminates adapters from the lower transformer layers during both training and inference, integrating the principles of these distinct approaches.
\citet{mahabadi2021parameter} illustrated the feasibility of learning adapter parameters across all layers and tasks by employing a shared hypernetwork, conditioned on task-specific and layer-specific details, to generate adapter parameters within the model, thus optimizing the fine-tuning process across various tasks.
\citet{DBLP:journals/corr/abs-2302-08106} proposed a reparameterizing the architecture, the general-purpose adaptive module can also be seamlessly integrated into most giant vision models, resulting in a zero cost in the inference process.
LoRA \cite{DBLP:conf/iclr/HuSWALWWC22} was a low-rank adaption that freezes the weights of pre-trained models and injects a trainable rank decomposition matrix into each layer of the Transformer architecture, thus significantly reducing the downstream number of trainable parameters for the task.

With the development and application of LLMs and visual models, researchers have realized that prompt has a large impact on the model's performance.
Prompt tuning \cite{lester2021power} only prepends and updates task-specific trainable parameters in the original input embeddings. 
\citet{DBLP:conf/cvpr/ChenYCZL23} proposed a visual prompt framework based on iterative label mapping, which automatically remaps source labels to target labels and incrementally improves the target task accuracy of visual prompts.
\citet{DBLP:conf/acl/ZakenGR22} proposed BitFit, a sparse fine-tuning method that modifies only the bias term of the model (or a subset of it). The delta fine-tuning \cite{DBLP:journals/corr/abs-2203-06904} fine-tunes only a small fraction of the model parameters while keeping the rest of the parameters unchanged, greatly reducing computational and storage costs.  
\citet{DBLP:conf/nips/LianZFW22} proposed a new parameter-efficient fine-tuning method, indicating that researchers can catch up with fully fine-tuned performance by simply scaling and shifting deep features extracted by the pre-trained model.

\subsection{Latent Space}
\citet{DBLP:conf/www/0001ZHZ022} is a universal latent embedding space for sentences that are first pre-trained on a large text corpus and then fine-tuned for various language generation and understanding tasks. \citet{DBLP:conf/emnlp/LiGLPLZG20} confirms the importance of learning latent space.  DISCODVT \cite{DBLP:conf/emnlp/JiH21} learns a latent variable sequence with each latent code abstracting a local text span to the discourse structures that guide the model to generate long texts with better long-range coherence.
There are also several studies \cite{liu-etal-2021-awakening, subramani-etal-2022-extracting} that incorporate latent structure learning into language model pre-training.


\section{Methodology}
To enhance the performance of PLMs in downstream tasks, it is essential to minimize the discrepancies between the representation obtained from the PLMs' encoder and the optimal input to the model decoder in the latent space. To this end, we propose calibrating the representation of PLMs in the latent space, as shown in Figure \ref{fig:model}. The output of the representation calibration block is directly added to the output of the PLM-based encoder to calibrate the input of the decoder. Our representation calibration block involves shape seed, learnable embedding and layer normalization. The details are as follows.

To calibrate the encoder's representation in the latent space, we introduce the concept of \(Shape~Seed\), a matrix designed to conform to the input's dimensions, facilitating the precise calibration of the encoder's representation.
The input of our representation calibration block is the \(Shape~Seed\) which is a matrix of size $batchsize \times n$. Here, $n$ equals the length of token embedding. We initialize the \(Shape~Seed\) to an all-ones matrix. 
Then, we use a learnable embedding layer $LearnEmb$ to encode \(Shape~Seed\) to obtain \(d_{i}\), which serves as the calibrated values.
Next, we add the calibrated values \(d_{i}\) to the encoder output \(h_{i}\) in the latent space, yielding the calibrated output \(p_{i}\).
This process realigns the encoder's output to a more reasonable representation in the latent space, thus making the PLM more adaptable to downstream tasks.

Given the input $X$, the above-mentioned calibration process can be formulated as
\begin{align}
\left\{ h_ { i } \right\} _ { i = 1 } ^ { n } &= \text{Encoder}( X ),\\
\left\{ d_ { i } \right\} _ { i = 1 } ^ { n}   &= LearnEmb(Shape\ Seed),\\
\left\{ p_ { i } \right\} _ { i = 1 } ^ { n } &=\left\{ h_ { i } +\lambda * d_{i} \right\} _ { i = 1 } ^ { n }, \\
\hat{y} = & \text{Decoder} ( y _ { < t }, \left\{ p _ { i } \right\} _ { i = 1 } ^ { n } ),
\end{align} 
where $\lambda$ is a hyperparameter that controls the degree of calibration. If our calibration block is not used, the output of the original PLM-based models  is  
\begin{align}
 \hat{y} = \text{Decoder} ( y _ { < t }, \left\{ h_ { i } \right\} _ { i = 1 } ^ { n } ).
\end{align}

The representation calibration block in our method is very simple and plug-and-play. Only the learnable embedding layer brings a marginal increase in the number of model parameters, which is analyzed in Table~\ref{para} in  Section \ref{stat}. When we integrate the proposed calibration method into the existing PLM-based models in the downstream tasks, it is unnecessary to change the loss function $\mathcal{L} _ {LM}$ used in these models.


\section{Experiments on Fine-tuning Methods}
\subsection{Tasks and Datasets}
We compare the proposed method with three fine-tuning methods using the SST-2, RET, MNLI,and CoLA datasets \cite{wang2018glue} to highlight the advantages and improvements.  The results of the baselines are from \citet{DBLP:journals/corr/abs-2203-06904}. We conduct experiments on 3 different random seeds, and \textbf{the reported results are the average of these three experiments.}

\begin{table*}[t]
\centering
\onehalfspacing
\resizebox{1.98\columnwidth}{!}{
\begin{tabular}{@{}l|c|c|c|c|c|c@{}}
\toprule[1.5pt]

\multirow{2}{*}{\textbf{Task \& Dataset}} & \multirow{2}{*}{Additional} 
& \multicolumn{3}{c}{Accuracy ($\uparrow$)}  
& \multicolumn{1}{c}{MCC ($\uparrow$)} 
& \multicolumn{1}{c}{AVERAGE ($\uparrow$)} \\
\cmidrule(lr){3-5} 

 & & SST2 & RET & MNLI & COLA &  \\

\midrule[1pt]
Prompt tuning \cite{lester2021power}  &0.03\% &92.20 &45.32 &35.43  &0.00 &43.23\\
\midrule[0.5pt]
Prefix-tuning  \cite{DBLP:conf/acl/LiL20} &7.93\% &92.66 &72.66 &82.21 &50.95 &74.62 \\
\midrule[0.5pt]
Adapter \cite{DBLP:conf/icml/HoulsbyGJMLGAG19} &2.38\% & 93.35 & 78.42 &83.90 &44.66 &75.08\\
\midrule[0.5pt]
LoRA \cite{DBLP:conf/iclr/HuSWALWWC22} & 0.38\% & 92.29 &79.14 &83.74 &49.40 &76.14 \\
\midrule[0.5pt]
BitFit $\heartsuit$  \cite{DBLP:conf/acl/ZakenGR22}   & 0.22\% & 93.20 &75.30 &84.10 &\textbf{53.20} &76.45\\
\midrule[0.5pt]
\textbf{RepCali}& 0.35\% & \textbf{94.31 }&\textbf{80.04} &\textbf{84.69} &51.15 &\textbf{77.55} \\
\bottomrule[1.5pt]
\end{tabular}
}
\caption{Overall test performance on SST2, RET and COLA. We evaluate all these fine-tuning methods on the T5-BASE backbone. The results of the baselines are from \citet{DBLP:journals/corr/abs-2203-06904}. $\heartsuit$ represents the results not from the original paper but reproduced by us.}
\label{sst}
\end{table*}

\subsection{Method Performance Comparison}
Our representation calibration method is a novel fine-tuning method. 
We mainly focus on tuning the latent representation from the PLMs' encoder; we froze the entire PLM decoder in NLU tasks to reduce the size of the fine-tuning parameters in RepCali while validating RepCali's calibration.

As shown in Table \ref{sst}, RepCali achieves the best results on all four tasks. Compared to LoRA, Adapter and Pefix-tuning, our method has over 1\% improvement on all four tasks. RepCali introduces only 0.35\% additional parameters to the T5-base model, and the increase is also less than the baseline mentioned above. This proves the simplicity and efficiency of our method, which requires only a small number of parameters to bring a huge improvement. The proposed RepCali is not only applicable to NLG tasks but equally used with NLU tasks. With RepCali, there is no need to consider where to add to the model, thus increasing the efficiency of fine-tuning.



\subsection{Comparison of Additional Parameters}
\label{gs_com}
As demonstrated in Table \ref{gs}, we quantify the additional parameters introduced by various fine-tuning approaches. Notably, our method contributes a minimal increase in parameter count, underscoring its efficiency in enhancing model performance without significantly expanding its complexity.

\begin{table*}[h]
\centering
\doublespacing
\scalebox{0.71}{
\begin{tabular}{l|l|l}
\toprule
\textbf{Name} & \textbf{Method} & \textbf{\#Params} \\
\midrule

Adapter \cite{DBLP:conf/icml/HoulsbyGJMLGAG19}  
& LayerNorm$(X + H(X)) \rightarrow$ LayerNorm$(X + ADT(H(X)))$ 
& $L \times 2 \times (2d_h d_m)$ \\
& $ADT(X) = X + \sigma(\mathbf{XW}_{d_h \times d_m}) \mathbf{W}_{d_m \times d_h},\ \sigma = \text{activation}$ 
& $(L - n) \times 2 \times (2d_h d_m)$ \\

\midrule

Prefix-tuning \cite{DBLP:conf/acl/LiL20}  
& $H_i = ATT(XW_q^{(i)}, [MLP_k^{(i)}(P_k') : XW_k^{(i)}], [MLP_v^{(i)}(P_v') : XW_v^{(i)}])$ 
& $n \times d_m + d_m^2$ \\
& $MLP^{(i)}(X) = \sigma(\mathbf{XW}_{d_m \times d_m})\mathbf{W}^{(i)}_{d_m \times d_h}$  
& $+ L \times 2 \times d_h d_m$ \\
& $\mathbf{P}' = \mathbf{W}_{n \times d_m}$ 
& \\

\midrule

LoRA \cite{DBLP:conf/iclr/HuSWALWWC22}  
& $ADT(X) = \mathbf{XW}_{d_h \times d_m} \mathbf{W}_{d_m \times d_h}$ 
& $L \times 2 \times (2d_h d_m)$ \\

\midrule

\textbf{Ours (RepCali)}  
& $Decoder(Encoder(X)) \rightarrow Decoder(Encoder(X)+d_i)$, where $d_i = RepCali(X)$  
& $2 \times d_h$ \\
& $RepCali(X) = \text{LayerNorm(Learnable Embedding(Shape Seed))}$  
&  \\

\bottomrule
\end{tabular}
}
\caption{
Comparison between different fine-tuning methods. [:] is the concatenation operation; $d_h$ means the hidden dimension of the transformer model; $d_m$ is the intermediate dimension between down projection and up projection, where $d_m$ is far smaller than $d_h$. Prefix-tuning adds a prefix of $n$ past key/value vectors.
}
\label{gs}
\end{table*}

\section{Experiments on Downstream Tasks}
\subsection{Downstream Tasks and Datasets}
We conduct comprehensive experiments on 8 downstream tasks: \textbf{End-to-end Response Generation}, \textbf{Abductive Commonsense Reasoning ($\alpha$NLG) },  \textbf{Task-Oriented Dialogue System}, \textbf{KG-to-Text},  \textbf{Abstractive Summarization}, \textbf{Dialogue Summarization}, \textbf{Dialogue Response Generation},  and \textbf{Order Sentences}. We integrate our representation calibration method on a total of 25 different PLM-based models, and all of them are based on fine-tuning.  For a fair comparison, we follow the other training parameters published in the original papers. We conduct experiments on 3 different random seeds, and the reported results are the average of the 3 experiments. \textbf{Due to the page limitation, the details of the benchmark datasets, implementation details  and  Experiments Results are reported in Appendix \ref{details}, Appendix \ref{dataset} and Appendix \ref{otherex}.}

\subsection{Analysis of Experimental Results}



\begin{table}[t]
\centering
\onehalfspacing 
\resizebox{0.95\columnwidth}{!}{
\begin{tabular}{@{}lcc@{}}
\toprule[1.5pt]
\textbf{End-to-end Response Generation} & \multicolumn{2}{c}{MultiWOZ} \\
\cmidrule(lr){2-3}
\textbf{Models} & Inf/Suc ($\uparrow$) & B-4/Com ($\uparrow$) \\
\midrule[1pt]
MinTL (T5-small) \cite{DBLP:conf/emnlp/LinMWF20} & 80.04/72.71 & 19.11/95.49 \\
\textbf{MinTL (T5-small)+RepCali} & \textbf{82.08/74.07} & \textbf{19.58/97.66} \\
\midrule
MinTL (T5-base) \cite{DBLP:conf/emnlp/LinMWF20} & 82.15/74.44 & 18.59/96.88 \\
\textbf{MinTL (T5-base)+RepCali} & \textbf{83.75/76.08} & \textbf{19.75/99.68} \\
\midrule
MinTL (BART-large) \cite{DBLP:conf/emnlp/LinMWF20} & 84.88/74.91 & 17.89/97.78 \\
\textbf{MinTL (BART-large)+RepCali} & \textbf{88.99/80.28} & \textbf{19.35/103.90} \\
\midrule
MinTL (T5-large) \cite{DBLP:conf/emnlp/LinMWF20} $\heartsuit$ & 79.68/71.27 & 19.55/95.03 \\
\textbf{MinTL (T5-large)+RepCali} & \textbf{81.68/73.57} & \textbf{19.61/97.24} \\
\midrule
MinTL (T5-3B) \cite{DBLP:conf/emnlp/LinMWF20} $\heartsuit$ & 78.48/66.87 & 14.65/87.33 \\
\textbf{MinTL (T5-3B)+RepCali} & \textbf{81.98/70.57} & \textbf{15.87/92.15} \\
\bottomrule[1.5pt]
\end{tabular}
}
\caption{End-to-end Response Generation results on MultiWOZ2. $\heartsuit$ indicates results reproduced by us; others are from the original paper.}
\label{e2e}
\end{table}

\begin{table}[t]
\centering
\onehalfspacing 
\resizebox{0.95\columnwidth}{!}{
\begin{tabular}{@{}lcc@{}}
\toprule[1.5pt]
$\boldsymbol{\alpha}$\textbf{NLG} & \multicolumn{2}{c}{$\mathcal{A R T}$} \\
\cmidrule(lr){2-3}
\textbf{Models} & SB-3/4 ($\downarrow$) & B-4/R-L ($\uparrow$) \\
\midrule[1pt]
BART-base \cite{DBLP:conf/acl/LewisLGGMLSZ20} & 56.32/52.44 & 13.53/38.42 \\
\textbf{BART-base+RepCali} & \textbf{48.13/49.24} & \textbf{14.42/39.66} \\
\midrule
MoE\_embed \cite{DBLP:conf/emnlp/ChoSH19} & 29.02/24.19 & 14.31/38.91 \\
\textbf{MoE\_embed+RepCali} & \textbf{29.01/23.92} & \textbf{14.90/39.71} \\
\midrule
MoE\_prompt \cite{DBLP:conf/icml/ShenOAR19} & 28.05/23.18 & 14.26/38.78 \\
\textbf{MoE\_prompt+RepCali} & \textbf{27.93/22.02} & \textbf{15.91/40.75} \\
\midrule
MoKGE \cite{DBLP:conf/acl/00020QZ0022} & 27.40/22.43 & 14.17/38.82 \\
\textbf{MoKGE+RepCali} & \textbf{24.67/19.07} & \textbf{15.25/40.16} \\
\bottomrule[1.5pt]
\end{tabular}
}
\caption{Diversity and quality evaluation on the $\alpha$NLG dataset. Baselines are from the original papers.}
\label{anlg}
\end{table}

\begin{table}[t]
\centering
\onehalfspacing 
\resizebox{0.95\columnwidth}{!}{
\begin{tabular}{@{}lcc@{}}
\toprule[1.5pt]
\textbf{Task-Oriented Dialogue System} & \multicolumn{2}{c}{CamRest} \\
\cmidrule(lr){2-3}
\textbf{Models} & BLEU-4 ($\uparrow$) & F1 ($\uparrow$) \\
\midrule[1pt]
BART-base \cite{DBLP:conf/acl/LewisLGGMLSZ20} & 19.05 & 55.92 \\
\textbf{BART-base+RepCali} & \textbf{19.56} & \textbf{56.40} \\
\midrule
T5-base \cite{DBLP:journals/jmlr/RaffelSRLNMZLL20} & 18.73 & 56.31 \\
\textbf{T5-base+RepCali} & \textbf{19.04} & \textbf{57.34} \\
\midrule
KB\_BART \cite{DBLP:journals/corr/abs-2201-08687} & 20.24 & 56.70 \\
\textbf{KB\_BART+RepCali} & \textbf{22.61} & \textbf{60.52} \\
\midrule
KB\_T5 \cite{DBLP:journals/corr/abs-2201-08687} & 21.11 & 59.67 \\
\textbf{KB\_T5+RepCali} & \textbf{21.78} & \textbf{61.78} \\
\midrule
KB\_T5 (large) \cite{DBLP:journals/corr/abs-2201-08687} $\heartsuit$ & 21.12 & 63.54 \\
\textbf{KB\_T5 (large)+RepCali} & \textbf{21.72} & \textbf{64.76} \\
\bottomrule[1.5pt]
\end{tabular}
}
\caption{Results on CamRest. $\heartsuit$ indicates results reproduced by us; others are from the original paper.}
\label{tod}
\end{table}

\begin{table}[t]
\centering
\onehalfspacing 
\resizebox{0.95\columnwidth}{!}{
\begin{tabular}{@{}lc@{}}
\toprule[1.5pt]
\textbf{KG-to-Text} & WebNLG \\
\cmidrule(lr){2-2}
\textbf{Models} & BLEU-4 / METEOR / R-L ($\uparrow$) \\
\midrule[1pt]
BART-base \cite{DBLP:conf/acl/LewisLGGMLSZ20} & 64.55 / 46.51 / 75.13 \\
\textbf{BART-base+RepCali} & \textbf{64.76 / 46.72 / 75.38} \\
\midrule
T5-base \cite{DBLP:journals/jmlr/RaffelSRLNMZLL20} & 64.42 / 46.58 / 74.77 \\
\textbf{T5-base+RepCali} & \textbf{64.90 / 46.83 / 75.14} \\
\midrule
JointGT (BART) \cite{DBLP:conf/acl/KeJRCWSZH21} & 65.92 / 47.15 / 76.10 \\
\textbf{JointGT (BART)+RepCali} & \textbf{66.10 / 47.35 / 76.18} \\
\midrule
JointGT (T5) \cite{DBLP:conf/acl/KeJRCWSZH21} & 66.14 / 47.25 / 75.90 \\
\textbf{JointGT (T5)+RepCali} & \textbf{66.72 / 47.46 / 76.46} \\
\midrule
GAP (BART) \cite{DBLP:conf/coling/ColasAW22} & \textbf{66.20} / 46.77 / 76.36 \\
\textbf{GAP (BART)+RepCali} & \textbf{66.20 / 46.89 / 76.41} \\
\bottomrule[1.5pt]
\end{tabular}
}
\caption{KG-to-Text results on WebNLG. Baselines are from the original paper.}
\label{kg3}
\end{table}

\begin{table}[t]
\centering
\onehalfspacing 
\resizebox{0.96\columnwidth}{!}{
\begin{tabular}{@{}lc@{}}
\toprule[1.5pt]
\textbf{Abstractive Summarization} & XSum \\
\cmidrule(lr){2-2}
\textbf{Models} & R-1 / R-2 / R-L ($\uparrow$) \\
\midrule[1pt]
BART-large \cite{DBLP:conf/acl/LewisLGGMLSZ20} & 45.14 / 22.27 / 37.25 \\
\textbf{BART-large+RepCali} & \textbf{45.42 / 22.60 / 37.63} \\
\midrule
PEGASUS \cite{DBLP:conf/icml/ZhangZSL20} & 47.46 / 24.69 / 39.53 \\
\textbf{PEGASUS+RepCali} & \textbf{47.78 / 24.75 / 39.70} \\
\midrule
BRIO-Mul \cite{DBLP:conf/acl/LiuLRN22} & 49.07 / 25.29 / 49.40 \\
\textbf{BRIO-Mul+RepCali} & \textbf{49.18 / 25.50 / 49.49} \\
\bottomrule[1.5pt]
\end{tabular}
}
\caption{Abstractive Summarization results on the XSum dataset.}
\label{sum}
\end{table}

\begin{table}[t]
\centering
\onehalfspacing 
\resizebox{0.95\columnwidth}{!}{
\begin{tabular}{@{}lc@{}}
\toprule[1.5pt]
\textbf{Dialogue Response Generation} & PersonaChat \\
\cmidrule(lr){2-2}
\textbf{Models} & R-1 / R-2 / R-L ($\uparrow$) \\
\midrule[1pt]
Blenderbot \cite{DBLP:conf/eacl/RollerDGJWLXOSB21} & 17.02 / 2.73 / 14.52 \\
\textbf{Blenderbot+RepCali} & \textbf{18.53 / 3.21 / 15.66} \\
\midrule
Keyword-Control \cite{DBLP:conf/acl/JiKGH22} & 17.31 / 3.02 / 14.81 \\
\textbf{Keyword-Control+RepCali} & \textbf{17.98 / 3.07 / 15.30} \\
\midrule
Focus-Vector \cite{DBLP:conf/acl/JiKGH22} & 20.81 / 3.98 / 17.58 \\
\textbf{Focus-Vector+RepCali} & \textbf{21.28 / 4.19 / 17.96} \\
\bottomrule[1.5pt]
\end{tabular}
}
\caption{Dialogue Response Generation results on PersonaChat. Baselines are from the original paper.}
\label{drg}
\end{table}

\textbf{End-to-End Response Generation:} 
We conduct a comprehensive evaluation of various models using the MultiWOZ dataset \cite{DBLP:conf/emnlp/BudzianowskiWTC18}. Within the MinTL framework \cite{DBLP:conf/emnlp/LinMWF20}, we incorporate our calibration method, encompassing BART-large \cite{DBLP:conf/acl/LewisLGGMLSZ20} and various sizes of T5 models (small, base, large, 3B) \cite{DBLP:journals/jmlr/RaffelSRLNMZLL20}. Following \citet{DBLP:conf/emnlp/LinMWF20}, we evaluate the models using metrics like Inform, Success, BLEU-4, and Combined((Inform+Success)$\times$0.5+BLEU-4) \cite{DBLP:conf/acl/PapineniRWZ02, DBLP:conf/sigdial/MehriSE19}.

As demonstrated in Table \ref{e2e}, the performance of each of the four baseline models exhibits significant enhancement following the incorporation of our representation calibration block. Notably, the MinTL(BART-large) demonstrates enhancements of 4.11\% in Inform, 5.37\% in Success, 1.46\% in BLEU-4, and 6.21\% in Combined score. Our method significantly enhances large language models (LLMs), underscoring their generality and effectiveness. We observe that in the MinTL framework, T5-3B's performance is lower than T5-base, potentially due to overfitting by the larger models (especially in small datasets) and hyperparameter settings.

\noindent\textbf{Abductive Commonsense Reasoning ($\alpha$NLG):}  We utilize the $\mathcal{A R T}$ benchmark dataset \cite{DBLP:conf/iclr/BhagavatulaBMSH20}, following the data split as \citet{DBLP:conf/acl/00020QZ0022}. We integrate our RepCali block   on BART-base \cite{DBLP:conf/acl/LewisLGGMLSZ20}, MoE-based methods \cite{DBLP:conf/emnlp/ChoSH19,DBLP:conf/icml/ShenOAR19}, MoKGE \cite{ DBLP:conf/acl/00020QZ0022}.
In line with \citet{DBLP:conf/acl/00020QZ0022}, we employ Self-BLEU3/4 \cite{DBLP:conf/sigir/ZhuLZGZWY18} as metrics of diversity assessment and BLEU-4 \cite{DBLP:conf/acl/PapineniRWZ02} and ROUGE-L\cite{lin2004rouge} as metrics of generation quality.

As shown in Table \ref{anlg}, by only employing our method, there are large improvements for all the baselines. For the previous SOTA model MoKGE, there is an improvement of 2.73\% and 3.36\% on the diversity metrics Self-BLEU-3/4 and an improvement of 1.08\% and 1.34\% on the quality of generation metrics BLEU-4 and ROUGE-L, respectively.  This proves that RepCali effectively makes the encoder's output more adaptable to the decoder.

\noindent\textbf{Task-Oriented Dialogue System:} We implement our RepCali block  on BART-base, T5-base, KB\_BART \cite{DBLP:journals/corr/abs-2201-08687}, and KB\_T5 \cite{DBLP:journals/corr/abs-2201-08687}. Following \citet{DBLP:journals/corr/abs-2201-08687}, we utilize the CamRest dataset \cite{wen2016network} and employ BLEU-4 and F1 scores.

As indicated in Table \ref{tod}, employing our representation calibration method led to a significant improvement in all four baseline models. Particularly for KB\_BART, it improved by 2.370\% and 3.818\% on BLEU-4 and F1 scores, respectively. 

\label{KG-to-Text}
\noindent\textbf{KG-to-Text:}  We implement our RepCali block   on BART-base \cite{DBLP:conf/acl/LewisLGGMLSZ20}, T5-base \cite{DBLP:journals/jmlr/RaffelSRLNMZLL20}, JointGT(BART) \cite{DBLP:conf/acl/KeJRCWSZH21}, JointGT(T5) \cite{DBLP:conf/acl/KeJRCWSZH21}, and GAP \cite{DBLP:conf/coling/ColasAW22}. Following \citet{DBLP:conf/acl/KeJRCWSZH21,DBLP:conf/coling/ColasAW22}, we utilize the WebNLG \cite{DBLP:conf/inlg/ShimorinaG18} dataset and employe BLEU-4, METEOR \cite{DBLP:conf/acl/BanerjeeL05}, and ROUGE-L as evaluation metrics.

As indicated in Table \ref{kg3}, there is a notable improvement across all five baseline models with the employment of our method. Compared to JointGT (T5) on the three metrics, there is an improvement of 0.58\%, 0.21\%, and 0.56\%, respectively. RepCali significantly enhanced the model compared to previous work. For example, compared to JointGT, GAP improved by 0.28\% and 0.26\% in BLEU-4 and R-L, respectively, but decreased by 0.38\% in METEOR. Whereas JointGT gets 0.18\%, 0.20\%, and 0.08\% improvement in the three metrics after using RepCali. This demonstrates that RepCali is a reasonable enhancement to the model, with improvements in all metrics. Compared to previous work, RepCali brings significant improvements by adding only a small number of parameters.

\noindent\textbf{Abstractive Summarization:} We conduct  Abstractive Summarization task using the XSum \cite{narayan-etal-2018-dont} dataset. We implement our RepCali block  on BART-large \cite{DBLP:conf/acl/LewisLGGMLSZ20}, PEGASUS \cite{DBLP:conf/icml/ZhangZSL20}, and BRIO \cite{DBLP:conf/acl/LiuLRN22}. In line with \citet{DBLP:conf/acl/LiuLRN22}, we employ ROUGE-1, ROUGE-2, and ROUGE-L \cite{lin2004rouge} as the evaluation metrics. 

As indicated in Table \ref{sum}, there is a notable enhancement in all three baseline models with the employment of our representation calibration block. For the Sota model BRIO-Mul, there is an improvement of 0.11\%, 0.21\% and 0.09\% on the three metrics, respectively. Although some of the metric improvement is minor, this improvement is significant compared to previous work.

\noindent\textbf{Dialogue Response Generation:} We conducte Dialogue Response Generation experiments on PersonaChat \cite{DBLP:conf/acl/KielaWZDUS18} dataset. We employ our representation calibration block on  Blenderbot   \cite{DBLP:conf/eacl/RollerDGJWLXOSB21}, Keyword-Control   \cite{DBLP:conf/acl/JiKGH22} and Focus-Vector \cite{DBLP:conf/acl/JiKGH22}. Following  \citet{DBLP:conf/acl/JiKGH22}, we utilize ROUGE-1, ROUGE-2, and ROUGE-L as evaluation metrics.

As shown in Table \ref{drg}, there is a large improvement in all baseline models. Relative to Blenderbot, there is an improvement of 1.51\%, 0.48\% and 1.14\% on the three metrics, respectively. It further proves the generalization and effectiveness of our representation calibration method.


\begin{table}[t]
\centering
\onehalfspacing 
\resizebox{0.95\columnwidth}{!}{
\begin{tabular}{@{}lc@{}}
\toprule[1.5pt]
\textbf{Order Sentences} & ROCStories \\
\cmidrule(lr){2-2}
\textbf{Models} & ACC / PMR / $\tau$ ($\uparrow$) \\
\midrule[1pt]
BART \cite{DBLP:conf/acl/LewisLGGMLSZ20} & 80.42 / 63.50 / 0.85 \\
\textbf{BART+RepCali} & \textbf{82.36 / 64.67 / 0.87} \\
\midrule
RE-BART \cite{chowdhury2021everything} & 90.78 / 81.88 / 0.94 \\
\textbf{RE-BART+RepCali} & \textbf{91.16 / 82.68 / 0.94} \\
\bottomrule[1.5pt]
\end{tabular}
}
\caption{Order Sentences results on ROCStories. The baseline results are from the original paper.}
\label{roc}
\end{table}

\noindent\textbf{Order Sentences:} We conducte Order Sentences experiments on ROCStories dataset. We employ our representation calibration block on  BART and RE-BART \cite{chowdhury2021everything}. Following  \cite{chowdhury2021everything}, we utilize Accuracy(ACC), Perfect Match Ratio (PMR), and Kendall’s Tau ($\tau$) as evaluation metrics. 

As shown in Table \ref{roc}, there is a large improvement in all baseline models. Relative to BART, there is an improvement of 1.96\%, 1.17\%, and 0.2 on the three metrics, respectively. For the previous Sota model RE-BART, there is an improvement of 0.38\%, 0.8\%  on the accuracy (ACC) and perfect Match Ratio (PMR). This improvement is significant compared to previous work.

\begin{table}[t]
\centering
\setstretch{1}
\resizebox{0.95\columnwidth}{!}{
\begin{tabular}{@{}lccc@{}}
\toprule[1.5pt]
\multirow{1}{*}{\textbf{Models}} &\multicolumn{1}{c}{\textbf{Size}} & \multicolumn{1}{c}{\textbf{Models}} & \multicolumn{1}{c}{\textbf{Size}}\\ 
\midrule[1pt]
  MinTL(T5-small)   \cite{DBLP:conf/emnlp/LinMWF20}   &  102M 
& \textbf +RepCali  &  102M \\
\midrule[1pt]
  BART-base    \cite{DBLP:conf/acl/LewisLGGMLSZ20}    &  139M 
& \textbf +RepCali  &  140M \\
\midrule[0.5pt]
  MoE\_embed  \cite{DBLP:conf/emnlp/ChoSH19}    &   139M 
& \textbf +RepCali &   140M \\
\midrule[0.5pt]
  MoE\_prompt   \cite{DBLP:conf/icml/ShenOAR19}   &   139M
& \textbf +RepCali &   140M \\
\midrule[0.5pt]
  KB\_BART \cite{DBLP:journals/corr/abs-2201-08687}    &  140M
& \textbf +RepCali &  140M \\
\midrule[0.5pt]
  MoKGE \cite{DBLP:conf/icml/ShenOAR19}  &  145M
& \textbf +RepCali &  146M \\
\midrule[0.5pt]
  JointGT(BART) \cite{DBLP:conf/acl/KeJRCWSZH21}   &  160M
& \textbf +RepCali &  161M \\
\midrule[0.5pt]
  T5-base  \cite{DBLP:journals/jmlr/RaffelSRLNMZLL20}    &  220M
& \textbf +RepCali &  221M \\
\midrule[0.5pt]
  KB\_T5 \cite{DBLP:journals/corr/abs-2201-08687}    &  222M
& \textbf +RepCali & 223M \\
\midrule[0.5pt]
  JointGT(T5) \cite{DBLP:conf/acl/KeJRCWSZH21}  &  265M  
& \textbf +RepCali &  265M   \\
\midrule[1pt]
  MinTL(T5-base)   \cite{DBLP:conf/emnlp/LinMWF20}   &  360M 
& \textbf +RepCali  &  361M \\
\midrule[0.5pt]
  Blenderbot     \cite{DBLP:conf/eacl/RollerDGJWLXOSB21}  &  364M
& \textbf +RepCali &  365M \\
\midrule[0.5pt]
  Keyword-Control   \cite{DBLP:conf/acl/JiKGH22}   &  364M
& \textbf +RepCali &   365M \\
\midrule[0.5pt]
  Focus-Vector \cite{DBLP:conf/acl/JiKGH22}    &   364M
& \textbf +RepCali &  365M \\
\midrule[0.5pt]
  BART-large   \cite{DBLP:conf/acl/LewisLGGMLSZ20}    &  400M
& \textbf +RepCali &  407M \\
\midrule[0.5pt]
  RE-BART   \cite{chowdhury2021everything}      &  400M
& \textbf +RepCali &  407M \\
\midrule[0.5pt]
  PEGASUS   \cite{DBLP:conf/icml/ZhangZSL20}    &  569M
& \textbf +RepCali &  569M \\
\midrule[0.5pt]
  BRIO-Mul \cite{DBLP:conf/acl/LiuLRN22}  &   569M
& \textbf +RepCali &  570M \\
\midrule[1pt]
  MinTL(BART-large)   \cite{DBLP:conf/emnlp/LinMWF20}   &  609M 
& \textbf +RepCali  &  610M \\
\midrule[0.5pt]
  T5-large  \cite{DBLP:journals/jmlr/RaffelSRLNMZLL20}     &  770M
& \textbf +RepCali &  770M \\
\midrule[0.5pt]
  MinTL(T5-large)   \cite{DBLP:conf/emnlp/LinMWF20}   &  1.17B
& \textbf +RepCali &  1.17B\\
\midrule[0.5pt]
  MinTL(T5-3B)   \cite{DBLP:conf/emnlp/LinMWF20}   &  4.5B
& \textbf +RepCali &  4.5B\\
\bottomrule[1.5pt]
\end{tabular}
}
\caption{Size of all baseline models before and after adding our calibration block. M: Millon, B: Billion
}
\label{para}
\end{table}

Experimental results demonstrate that our representation calibration method offers desirable enhancements to PLMs (including LLMs) and significantly improves the performance of tasks. Experimental results on English and Chinese datasets show that RepCali can generalize to different languages effectively. This underscores the effectiveness and broad applicability of our representation calibration method. By minimizing the discrepancies between the representation obtained from the PLMs' encoder and the optimal input to the decoder of the model in the latent space, our method notably enhances the performance of the PLMs in downstream tasks. Remarkably, our method is both efficient and lightweight, involving only an additional learnable embedding layer. Despite its minimal impact on the model’s parameter count, it results in substantial performance improvements. Overall, RepCali adds only 0-0.8\% extra parameters yet delivers significant performance gains.

\begin{figure*}[t]
\centering
\includegraphics[width=1.96\columnwidth]{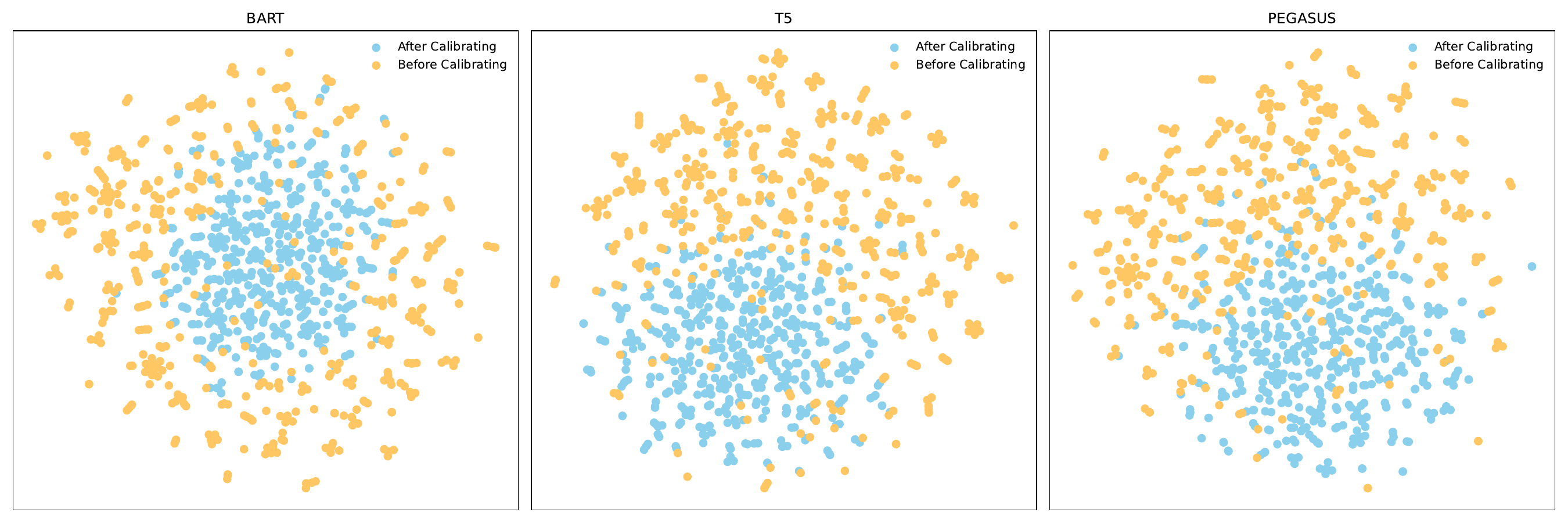}
\caption{ We chose three different PLMs, BART, T5, and PEGASUS for visualization and analysis of the latent space. The blue points are the hidden representations obtained after fine-tuning using our calibration method RepCali, and the yellow points are the hidden representations obtained after fine-tuning for the PLMs.}
\label{token}
\end{figure*}

\subsection{Model Sizes}
\label{stat}
As detailed in Table \ref{para}, the parameter growth for each model varies based on its hidden state dimension, e.g., BART-base has a hidden state dimension of 768 and BART-large has a hidden state dimension of 1024.  It reveals that the largest model, MinTL(T5-3B), boasts a formidable 4.5 billion parameters. This observation highlights the compatibility of our representation calibration method with large language models, consistently delivering valuable enhancements in LLMs.  We calculate the parametric quantities since they are not mentioned in the corresponding papers. Overall, our method only adds 0-0.8\% additional parameters.

\subsection{Visual Analysis in the Latent Space}
We use tSNE \cite{van2008visualizing} to visualize the learned feature on a 2D map. The validation set of Abductive Commonsense Reasoning ($\alpha$NLG) is used to extract the latent features. We chose three different PLMs, BART, T5, and PEGASUS for visualization and analysis of the latent space.  As shown in Figure \ref{token}, compared to the PLMs without RepCali for representation calibration, the PLMs with RepCali learn a smoother space with more organized latent patterns, while the latent representation is more compact, which is why the better performance of the boosted model can be obtained with RepCali. This coincides with the argument in works \cite{DBLP:conf/emnlp/LiGLPLZG20}, which suggests that smooth regularization on the latent space benefits the model's performance.

\section{Conclusion}
In this paper, we propose a generalized representation calibration method (RepCali) to minimize discrepancies between the representation obtained from the PLMs' encoder and the optimal input to the decoder of the model. During the fine-tuning phase, we integrate our representation calibration block to the latent space after the encoder and use the calibrated output as the decoder input. Our representation calibration method is suitable for all PLMs with encoder-decoder architectures, as well as the models based on PLMs. Our representation calibration method is both plug-and-play and easy to implement. Comparison experiments across
4 benchmark tasks indicate that RepCali is superior to the representative fine-tuning baselines.  Extensive experiments on 25 PLM-based models across 8 downstream tasks (including both English and Chinese datasets)  demonstrate that the proposed RepCali offers desirable enhancements to PLMs (including LLMs) and significantly improves the performance of downstream tasks. 




\bibliography{custom}

\appendix

\section{Implementation Details}
\label{details}
When applied to the downstream task, we integrate our representation calibration block on a total of 25 different PLM-based models (including LLM), all of which are based on fine-tuning. We conduct experiments on 3 different random seeds, and the reported results are the average of the 3 experiments. The baseline models used are full-model fine-tuned on the downstream tasks, and we also full-model fine-tuned after integrating our representation calibration block into the baseline models. For a fair comparison, we follow the other training parameters published in the original papers.

MinTL (T5-3B) experiments are performed on the NVIDIA Ampere A100 GPU, which boasts 80GB of memory. The remaining experiments use NVIDIA Pascal P40 GPUs with 24GB memory and NVIDIA V100 GPUs with 32GB memory.


\section{Experiments Datasets}
\label{dataset}
\textbf{End-to-End Response Generation:} We evaluate the models on the MultiWOZ dataset \cite{DBLP:conf/emnlp/BudzianowskiWTC18}. It is a large-scale multidomain task-oriented dialogue benchmark collected via the Wizard-of-Oz setting. The dataset
contains 8438/1000/1000 dialogues for training/validation/testing, respectively. 

\begin{table*}[t]
\centering
\resizebox{1.95\columnwidth}{!}{
\begin{tabular}{@{}lc|c|c|c|c|c@{}}
\toprule[1.5pt]
\textbf{Dialogue Summarization} & \multicolumn{6}{c}{CSDS (Chinese Dataset)} \\
\cmidrule(lr){2-7}
\textbf{User Summarization} & ROUGE-1 & ROUGE-2 & ROUGE-L & BLEU-4 & BERTScore & MoverScore \\
\midrule[1pt]
BART-base \cite{DBLP:conf/acl/LewisLGGMLSZ20} & 58.75 & 43.59 & 56.86 & 34.26 & 80.67 & 59.86 \\
\textbf{BART-base+RepCali} & \textbf{59.17} & \textbf{44.21} & \textbf{57.31} & \textbf{35.22} & \textbf{81.46} & \textbf{59.97} \\
\midrule
BART-both \cite{lin-etal-2022-roles} & 58.93 & 43.69 & 57.28 & 34.49 & 80.64 & 59.86 \\
\textbf{BART-both+RepCali} & \textbf{59.19} & \textbf{44.26} & \textbf{57.40} & \textbf{35.17} & \textbf{81.58} & \textbf{60.04} \\
\midrule
BART-GLC \cite{liang2023enhancing} & 61.42 & 45.83 & 59.25 & 36.43 & 81.83 & 61.03 \\
\textbf{BART-GLC+RepCali} & \textbf{61.80} & \textbf{46.05} & \textbf{59.60} & \textbf{36.76} & \textbf{82.47} & \textbf{61.13} \\
\midrule[1pt]
\textbf{Agent Summarization} & ROUGE-1 & ROUGE-2 & ROUGE-L & BLEU-4 & BERTScore & MoverScore \\
\midrule
BART-base \cite{DBLP:conf/acl/LewisLGGMLSZ20} & 53.89 & 40.24 & 50.85 & 31.88 & 77.31 & 58.75 \\
\textbf{BART-base+RepCali} & \textbf{54.05} & \textbf{40.37} & \textbf{50.94} & \textbf{32.11} & \textbf{77.73} & \textbf{58.83} \\
\midrule
BART-both \cite{lin-etal-2022-roles} & 54.01 & 40.32 & 51.10 & 32.30 & 77.30 & 58.73 \\
\textbf{BART-both+RepCali} & \textbf{54.12} & \textbf{40.34} & \textbf{51.14} & \textbf{32.47} & \textbf{78.03} & \textbf{58.96} \\
\midrule
BART-GLC \cite{liang2023enhancing} & 54.59 & 40.02 & 52.43 & \textbf{32.58} & 77.61 & 59.02 \\
\textbf{BART-GLC+RepCali} & \textbf{54.70} & \textbf{40.06} & \textbf{52.46} & 32.45 & \textbf{78.83} & \textbf{59.16} \\
\bottomrule[1.5pt]
\end{tabular}
}
\caption{Evaluation results for dialogue summarization on the CSDS dataset. Baseline results are from the original papers.}
\label{ds}
\end{table*}

\noindent\textbf{Diversity Abductive Commonsense Reasoning ($\alpha$NLG):} We use the $\mathcal{A R T}$  benchmark dataset \cite{DBLP:conf/iclr/BhagavatulaBMSH20} that consists of 50,481 / 1,779 / 3,560 examples for training/validation/validation sets. The average input/output length is 17.4 / 10.8 words. Each example in the $\mathcal{A R T}$ dataset has 1 to 5 references.

\noindent\textbf{Task-Oriented Dialogue System:} We use CamRest dataset \cite{wen2016network}, a human-to-human dialogues dataset for restaurant recommendation in Cambridge. 676 dialogues are provided by the CamRest dataset. It is split into 406, 135, and 135 as
training data, validation data, and test data, respectively. The
templates are generated from training data and augment 9,728
new dialogues to the training data.

\noindent\textbf{Abstractive Summarization:} XSum  \cite{narayan-etal-2018-dont} is a highly abstractive
dataset of articles from the British Broadcasting
Corporation (BBC). Xsum consists of  203K/11k/11k examples for training/development/test sets.

\noindent\textbf{KG-to-Text:} WebNLG is a crowd-sourced RDF triple-to-text dataset manually crafted by human annotators.
The dataset contains graphs from DBpedia \cite{DBLP:conf/semweb/AuerBKLCI07} with up to 7 triples paired with one or more reference texts. It consists of  34352/4316/4224 examples for training/validation/testing sets.

\noindent\textbf{Dialogue Response Generation:}
For the Dialogue Response
Generation task, we adopt the PersonaChat
dataset \cite{DBLP:conf/acl/KielaWZDUS18}. It is an open-domain
multi-turn chit-chat dataset, where two participants
are required to get to know each other by chatting
naturally. The PersonaChat dataset contains 8,939 dialogues
for training, 1,000 for validation, and 968 for testing.
For each turn in the dialogue, we concatenate the persona of the speaker and the dialogue history as input and train the base model to generate the current utterance.

\noindent\textbf{Order Sentences:}  We randomly split ROCStories into
train/test/validation in an 80:10:10 ratio. For
the other datasets, we use the same train, test, and
validation sets as previous works.

\noindent\textbf{Dialogue Summarization:}  CSDS is the first role-oriented dialogue summarization Chinese dataset, which provides separate summaries for users and agents (customer service). The CSDS dataset contains 9101 dialogues for training, 800 for validation, and 800 for testing.

\label{resutls}
\label{kg2}

\section{Other Experiments Results}
\label{otherex}
\noindent\textbf{Dialogue Summarization: }
We conducte Dialogue Summarization experiments on the CSDS \cite{lin-etal-2021-csds} dataset. CSDS is the first role-oriented dialogue summarization Chinese dataset, which provides separate summaries
for users and agents (customer service). We employ our representation calibration block on  BART and GLC \cite{liang2023enhancing}. Following  \citet{liang2023enhancing}, we utilize ROUGE-1, ROUGE-2, ROUGE-L, Bleu-4, BERTScore and MoverScore as evaluation metrics.

As indicated in Table \ref{ds}, there is a notable improvement across all baseline models with the employment of our method. Regarding user summaries, the six metrics improved by 0.38\%, 0.22\%, 0.35\%, 0.33\%, 0.64\%, and 0.10\%, respectively, compared to the SOTA model, BART-GLC.
Regarding Agent summarization, there is an improvement of 0.11\%, 0.04\%, 0.03\%, 1.22\%, and 0.14\% at ROUGE-1, ROUGE-2, ROUGE-L, BERTScore, and MoverScore, respectively, when compared to the SOTA model BART-GLC. Although some of the metric improvement is minor, this improvement is significant compared to previous work. The significant improvement in BERTScore suggests that the text generated after using RepCali is more semantically logical and coherent.

\end{document}